%% file: iclr2027_conference.tex
\documentclass{article} 
\usepackage{iclr2027_conference,times}

\input{math_commands.tex}

\usepackage{hyperref}
\usepackage{url}

\usepackage{graphicx} 
\usepackage{booktabs}
\usepackage{float}

\usepackage{titletoc}

\title{The TIME Machine: On The Power of Motion for Efficient Perception}

\author{%
  Mantas Skackauskas\thanks{Correspondence to: \texttt{m.skackauskas@sms.ed.ac.uk}} \quad Xinyue Hao \quad Laura Sevilla-Lara \\
  School of Informatics\\
  University of Edinburgh\\
}
\usepackage[table]{xcolor}
\definecolor{lightblue}{RGB}{230,245,255}

\newcommand{\vjepa}{V-JEPA 2}
\newcommand{\emb}{TIME}

\newcommand{\sota}{state-of-the-art}

\iclrfinalcopy 

\begin{document}

\maketitle

\input{abstract}
\input{introduction}

\input{related_work}
\input{methods}

\input{experimental_results}

\input{limitations}
\input{conclusion}

\break

\bibliography{iclr2027_conference}
\bibliographystyle{iclr2027_conference}

\clearpage
\section{Appendix}

In this supplementary material, we include the following:

\appendix

\startcontents[appendix]

\titlecontents{section}[0pt]
  {\bfseries}
  {\makebox[1.5em][l]{\thecontentslabel}}
  {}
  {}

\titlecontents{subsection}[1.5em]
  {}
  {\makebox[2.0em][l]{\thecontentslabel}}
  {}
  {}    

\printcontents[appendix]{}{0}{}

\break
\section{Additional Experimental Results}

\label{appendix:appendix_additional_results}
\subsection{Motion-only baseline comparison}
\label{appendix:motion_only_baseline_results_comparison}

As a sanity check, we include a comparison via a linear probe between our \emb~ model representations and the corresponding backbone features of CoTracker3 \citep{cotracker3}, TAPIR \citep{doersch2023tapirtrackingpointperframe}, and features obtained from raw trajectories. Our model maintains the lead in all benchmarks. Results can be seen in Table~\ref{tab:motion_only_baseline_results}.
\input{tables/motion_only_baseline_results}

\subsection{Video Understanding Task Standalone Model Performance}

\label{appendix:in_the_wild_appendix}
Full SSv2, Diving48 and Ego-Exo4D linear probing results including standalone model performance can be found in Table \ref{tab:all_probing_results_appendix}.

\begin{table}[h]
\centering
\caption{\textbf{Video Understanding Task Performance.} We report top-1 accuracy. Here, $\Delta$ denotes the improvement over the baseline. \\}
\label{tab:all_probing_results_appendix}
\resizebox{\textwidth}{!}{%
\begin{tabular}{@{}l cc cc cc@{}}
\toprule
& \multicolumn{2}{c}{\textbf{Ego-Exo4D}} & \multicolumn{2}{c}{\textbf{SSv2}} & \multicolumn{2}{c}{\textbf{Diving48}} \\
\cmidrule(lr){2-3} \cmidrule(lr){4-5} \cmidrule(lr){6-7}
\textbf{Model} & \textbf{Top-1 ($\pm$ Std \%)} & \textbf{$\Delta$} & \textbf{Top-1 ($\pm$ Std \%)} & \textbf{$\Delta$} & \textbf{Top-1 ($\pm$ Std \%)} & \textbf{$\Delta$} \\
\midrule
\multicolumn{7}{l}{\textit{Image Foundation Models}} \\
CLIP (4f) \citep{clip} & 25.11 $\pm$ 0.19 & -- & 18.56 $\pm$ 0.04 & -- & 13.97 $\pm$ 0.25 & -- \\
DINOv3 (4f) \citep{dino} & 24.88 $\pm$ 0.13 & -- & 21.95 $\pm$ 0.06 & -- & 16.35 $\pm$ 0.19 & -- \\
\midrule
\multicolumn{7}{l}{\textit{Point Trajectory Encoder}} \\
TIME (Point Tracks) & 17.27 $\pm$ 0.12 & -- & 19.42 $\pm$ 0.04 & -- & 8.37 $\pm$ 0.15 & -- \\
\midrule
\multicolumn{7}{l}{\textit{Video Foundation Models}} \\
RVM$^\dagger$ \citep{recurrentvideo} & 25.67 $\pm$ 0.18 & -- & 40.08 $\pm$ 0.07 & -- & 9.21 $\pm$ 0.16 & -- \\
VideoMAEv2$^\dagger$ \citep{wang2023videomaev2scalingvideo} & 41.01 $\pm$ 0.15 & -- & 33.05 $\pm$ 0.07 & -- & 24.24 $\pm$ 0.42 & -- \\
V-JEPA 2$^\dagger$ \citep{vjepa2} & 34.85 $\pm$ 0.24 & -- & 49.78 $\pm$ 0.05 & -- & 30.08 $\pm$ 0.30 & -- \\
\midrule
\multicolumn{7}{l}{\textbf{\textit{Ours + Spatial / Spatiotemporal}}} \\
TIME + CLIP (4f) & 39.33 $\pm$ 0.24 & \textcolor{green!60!black}{$\mathbf{+14.22}$} & 31.56 $\pm$ 0.05 & \textcolor{green!60!black}{$\mathbf{+13.00}$} & 16.11 $\pm$ 0.28 & \textcolor{green!60!black}{$\mathbf{+2.14}$} \\
TIME + DINOv3 (4f) & 39.63 $\pm$ 0.22 & \textcolor{green!60!black}{$\mathbf{+14.75}$} & 34.60 $\pm$ 0.10 & \textcolor{green!60!black}{$\mathbf{+12.65}$} & 18.08 $\pm$ 0.21 & \textcolor{green!60!black}{$\mathbf{+1.73}$} \\
TIME + RVM$^\dagger$ & 37.00 $\pm$ 0.17 & \textcolor{green!60!black}{$\mathbf{+11.32}$} & 41.92 $\pm$ 0.06 & \textcolor{green!60!black}{$\mathbf{+1.84}$} & 11.74 $\pm$ 0.11 & \textcolor{green!60!black}{$\mathbf{+2.53}$} \\
TIME + VideoMAEv2$^\dagger$ & 51.14 $\pm$ 0.23 & \textcolor{green!60!black}{$\mathbf{+10.12}$} & 38.74 $\pm$ 0.09 & \textcolor{green!60!black}{$\mathbf{+5.69}$} & 24.97 $\pm$ 0.24 & \textcolor{green!60!black}{$\mathbf{+0.73}$} \\
TIME + V-JEPA 2$^\dagger$ & 45.60 $\pm$ 0.12 & \textcolor{green!60!black}{$\mathbf{+10.75}$} & 50.00 $\pm$ 0.10 & \textcolor{green!60!black}{$\mathbf{+0.22}$} & 30.04 $\pm$ 0.15 & \textcolor{gray}{$\mathbf{-0.04}$} \\
\bottomrule
\multicolumn{7}{l}{\footnotesize $^\dagger$ Denotes models whose pre-training data explicitly includes the SSv2 dataset.} \\
\end{tabular}%
}
\end{table}

\subsection{Attentive Probing}

\label{appendix:attentive_probe}

We experiment with the attentive probing protocol \citep{bardes2024revisitingfeaturepredictionlearning}. Due to hardware resource constraints (e.g., storing raw unpooled video model encoder features required for training the attentive probe needs terabytes of storage), we follow the low-shot frozen feature evaluation protocol from the V-JEPA v1 paper \citep{bardes2024revisitingfeaturepredictionlearning}. We extract and store raw features from each model (TIME, V-JEPA 2, RVM, VideoMAEv2) from 5\% of the training data ($\approx48$ samples per class). We then follow the attentive probe protocol from the publicly available code \citep{carreira2025scaling4drepresentations} and train the attentive probes on the extracted frozen training data features. We evaluate on the frozen validation set features. We repeat the experiment 3 times and report the mean accuracy and the standard deviation. We confirm the correctness of the attentive probing protocol based on VideoMAEv2 performance in the original V-JEPA v1 paper \citep{bardes2024revisitingfeaturepredictionlearning}. In this work, the reported accuracy of VideoMAEv2 $5\%$ low-shot attentive probe ($28.0 \pm 1.0$) aligns with that of our protocol implementation ($28.22 \pm 0.96$) in Table \ref{tab:attentive_probe_results_table}. We conclude that, under the attentive probe protocol,~\emb~ benefits all models.

\begin{table}[h]
\centering
\caption{\textbf{Low-Shot (5\%) Attentive Probe Evaluation on SSv2.} For concatenated models, $\Delta$ indicates the improvement over the corresponding standalone model.\\}
\label{tab:attentive_probe_results_table}
\begin{tabular}{@{}lcc@{}}
\toprule
\textbf{Model} & \textbf{SSv2 Acc. (\%)} & \textbf{$\Delta$} \\
\midrule
TIME (Ours) & 12.92 $\pm$ 1.02 & -- \\
VideoMAEv2 \citep{wang2023videomaev2scalingvideo} & 28.22 $\pm$ 0.96 & -- \\
TIME + VideoMAEv2 & 29.46 $\pm$ 0.83 & \textcolor{green!60!black}{$\mathbf{+1.24}$} \\
RVM \citep{recurrentvideo} & 39.83 $\pm$ 0.71 & -- \\
TIME + RVM & 42.16 $\pm$ 0.78 & \textcolor{green!60!black}{$\mathbf{+2.33}$} \\
V-JEPA 2 \citep{vjepa2} & 48.96 $\pm$ 0.43 & -- \\
TIME + V-JEPA 2 & 50.21 $\pm$ 0.41 & \textcolor{green!60!black}{$\mathbf{+1.25}$} \\
\bottomrule
\end{tabular}
\end{table}





\section{Additional Ablation Studies}
\label{appendix:additional_ablation_studies}

\subsection{Masking Strategy}
We explore how TIME deals with point trajectories that appear later, by performing a middle frame tracking experiment. We initialize the point trajectory grid from the middle frame and obtain trajectory paths in both directions. This enables us to place point trajectories on all objects that appear in the middle of the video rather than at the beginning. We switch to middle-frame trajectories directly at inference (i.e., using the original TIME model trained on first-frame trajectories). We train a linear probe on the full SSv2 dataset and report mean and standard deviation after 5 runs in Table~\ref{tab:masking_strategy_table}. As seen in the results, tracking from the middle frame leads to an additional $1\%$ gain in TIME linear probe performance. This suggests that the model performance may increase during the point grid initialization at a timestep when objects of interest are visible in the scene. However, while in SSv2 we do observe that objects of interest often appear after the first frame, this behavior is dataset-dependent. Therefore, determining the optimal time to initialize general point tracking is beyond the scope of this project, hence, we leave it for future work. Additionally, results suggest that the proposed model is invariant to the grid initialization time. 

\input{tables/masking_strategy_table}

\subsection{Sim-to-Real Generalization}
We further evaluate synthetic versus real-world trajectory impact on model pre-training. We train two~\emb~models from scratch. The first model is trained on a subset (50k) of Kubric simulations. The second model is trained on a subset (50k) of SSv2 trajectories obtained using CoTracker3. We evaluate these models on SSv2 validation set via a standard linear probe. Each experiment is repeated 5 times and the results are shown in the Table~\ref{tab:sim_to_real_generalization_table}. We observe that training on real-world trajectories leads to a minor increase in performance ($1.68\%$ relative gain) on the SSv2 validation set. Similarly to our ablation study in Figure \ref{fig:ablation}, the results suggest that synthetic trajectories already generalize well to real-world scenarios.

\input{tables/sim_to_real_generalization_table}


\subsection{Static Camera Simulations}
The synthetic training data pipeline produces samples with a static camera. During evaluation, however, \emb~does not assume a static camera as it is tested on datasets that include diverse camera motion, such as the full Something-Something v2 (SSV2) \citep{ssv2} dataset and the full Diving48 \citep{diving48} dataset. We conduct experiments to show that the proposed model is able to generalize well: for example some of the classes in the SSV2 datasets are types of camera motion, and the proposed method does particularly well in those. The performance of standalone TIME on action labels from SSV2 that are specific to camera motion can be seen in Table~\ref{tab:static_camera_results}. 

\input{tables/static_camera_results}

\subsection{Token Components}
We complete a leave-one-out ablation study by excluding KNN (max-deviation), global displacement and occlusion signals.The~\emb~models are trained from scratch on a subset (50k) of Kubric simulations and evaluated on SSv2 via a linear probe. Each experiment is repeated 5 times and the results are shown in Table~\ref{tab:token_information_ablation_table}. We report that each ablated component significantly impacts the encoder downstream task performance. Occlusion ($-23.66 \%$ drop) is the most impactful followed by neighborhood ($-20.90 \%$ drop) and global displacement ($-11.29 \%$ drop)  signals.

\input{tables/token_information_ablation_table}

\section{Method Design Choices}
\label{appendix:further_design_decisions}

\subsection{Zero-Shot Evaluation Setting}
Our reported evaluation is on zero-shot model performance from frozen features. Therefore, the models themselves are not pretrained on the evaluation datasets unless specified. As mentioned in the main paper, we follow the usual linear probe evaluation protocol from frozen features. First, we obtain training data and validation data features using a forward pass for each frozen model separately. Next, frozen training data features are then used to train linear probes. Finally, the corresponding frozen validation data features are passed to the trained linear probes to obtain the performance metrics. As a result, the evaluation on all models is zero-shot as only the linear probe is trained and evaluated on frozen model features.

\subsection{Late Fusion Choice}
We concatenate frozen features of TIME and standard video and image representations, and use a linear classifier on the full benchmarks. The goal of this work is to explore a novel temporal paradigm “without bells and whistles”, not to achieve the best possible results on the benchmarks. Hence the choice of fusion is indeed deliberately simple. Therefore, we leave the integration with images, languages, etc. for future works. A parallel work \citep{trackMAE}, attempts to fuse trajectories with RGB video during training. In contrast, our work specifically chooses to model motion independently of appearance. This enables us to test whether removing high-dimensionality associated with visual cues would help to learn temporal motion manifold more efficiently. As seen in Appendix~\ref{appendix:in_the_wild_appendix}, Table~\ref{tab:all_probing_results_appendix}, late fusion of TIME features with sparse spatial features from DINOv3 (4 video frames) allows us to exceed baseline VideoMAEv2 performance ($34.60 \pm 0.10$ versus $33.05 \pm 0.07$ respectively). This suggests that late fusion of spatial and temporal features may be a possible alternative to resource-intensive modelling of spatio-temporal representations directly.

\subsection{Loss Function Design}
\label{appendix:loss_clarification}
The original MAE reconstruction methodology considers all masked patches equally \citep{videomae}. As a result, we follow this methodology and treat each trajectory point (occluded or not) equally in the reconstruction objective as well. In addition, as we train purely on synthetic data, all point trajectories are ground-truth. This reliability component further forces the model to learn semantically rich events without differentiating between occlusion flags. Furthermore, deliberately excluding occlusion in the reconstruction objective lessens the effect of a distribution shift between occluded ground-truth points during training and noisy points during real-world inference tasks. Similarly, we keep the reconstruction loss only on local and global displacement tokens to maintain simplicity. We empirically find that this is sufficient to prevent mode-collapse during training, thus, we omit any additional token information during this step.

\subsection{Tokenization Details}
\label{appendix:tokenization_details}

In the proposed~\emb~model, corresponding tokens are organized along temporal and spatial axes:

\paragraph{Temporal axis.} Following the original VideoMAEv2 paper \citep{wang2023videomaev2scalingvideo}, we set the tubelet size to 2 consecutive frames. Thus, we group each of the 24 input frames and partition them into 12 non-overlapping tubelets maintaining the original temporal order.

\paragraph{Spatial axis.} Moving point trajectories are likely to diverge from their initialized 2D grid location over time. Therefore, we treat them as a flat, orderless set of 1024 points. Next, to model point location along the spatial axis, we assign each point a single learnable embedding vector. We repeat this across all 12 tubelets per each point. This mechanism enables the model to learn the spatial location of each point during training.

\section{\emb~ Model Training Details}
\subsection{Kubric pipeline settings}
We provide additional Kubric MOVi-B pipeline \citep{kubric} settings for experiment reproducability purposes in the Table~\ref{tab:kubric_settings} below.
\label{appendix:movi_pipeline_settings}

\begin{table}[h]
\centering
\caption{\textbf{Kubric dataset pipeline settings.}\\}
\label{tab:kubric_settings}
\begin{tabular}{@{}p{0.28\textwidth} p{0.65\textwidth}@{}}
\toprule
\textbf{Parameter} & \textbf{Distribution} \\
\midrule
Background & Solid floor \\
\# Objects & Uniform selection between 3 and 10 objects \\
Shapes & Uniform selection from 11 KuBasic assets: cube, cylinder, sphere, cone, torus, gear, torus knot, sponge, spot, teapot, suzanne \\
Scale & Uniform number between 0.7 and 1.4 (inclusive) \\
Material & Uniform choice between metal or rubber properties \\
Spawn region & Between regions $x,y \in [-5,5]$ and $z \in [1,5]$ without object overlap \\
Initial velocity & Uniform choice of values between $[-4,4]$ for $x$ and $y$ axes. Biased toward origin \\
Physics & PyBullet simulation for 24 frames at 12\,fps (2 seconds) \\
\bottomrule
\end{tabular}
\end{table}

\subsection{Kubric Occlusion Handling}
\label{appendix:occlusion_handling}
Synthetic trajectories generated by Kubric are marked as occluded if they are covered by another object or if they fall outside the image frame margins. The ``noise from occlusions'' line refers to the noisy real-world tracks obtained using point tracking. Using synthetic data, we can guarantee that occluded points are as accurate as non-occluded, thus preventing the proposed model from learning from incorrect point trajectories during training.

\subsection{Model training settings}
\label{appendix:full_time_pretraining_details_appendix}

In this section we provide more extensive details of our model training. We report the numbers that were used to train the \emb~ model on $250,000$ synthetic Kubric MOVi-B samples. More information regarding hyperparameters can be found in Table \ref{tab:full_time_hyperparam}. 

To repeat all the evaluation experiments (including feature extraction from pre-trained video model baselines and use of tracking software) in this paper, we estimate it would take \textless 100 GPU hours.

\begin{table}[htbp]
\centering
\caption{\textbf{\emb~ Pre-training Hyperparameters.} Our 250K model is trained for 100 epochs. Training on 4$\times$A100 80GB GPUs lasts $\approx5$ days ($\approx$700 GPU hours).\\}
\label{tab:full_time_hyperparam}
\label{tab:trajectory_mae_pretrain}
\begin{tabular}{l c}
\toprule
\textbf{Parameter} & \textbf{Value} \\
\midrule
Number of frames & 24 \\
Frames per Second & 12.0 \\
Crop Size & 256 \\
Sampling Rate & 1 \\
\midrule
Grid Size (track points) & $32 \times 32$ \\
Number of Track Points & 1024 \\
Tubelet Size & 2 \\
Token Count per Sample & 12,288 \\
Mask Ratio & 0.75 \\
Visible Tokens & $\sim$1,230 (25\%) \\
\midrule
Encoder Depth & 12 \\
Decoder Depth & 4 \\
$k$-Nearest Neighbors & 16 \\
Loss Type & Huber ($\beta = 0.5$) \\
Motion Loss Boost & 7.0 \\
\midrule
Epochs & 100 \\
Warmup Epochs & 3 \\
Batch Size (per GPU) & 10 \\
Batch Size (global, 4$\times$GPUs) & 40 \\
Starting Learning Rate & $1.5 \times 10^{-4}$ \\
Final Learning Rate & $1.0 \times 10^{-6}$ \\
Weight Decay & 0.05 \\
Drop Path & 0.1 \\
Clip Gradient & 0.02 \\
\midrule
Optimizer & AdamW \\
Precision & AMP (Automatic Mixed Precision) \\
\midrule
Data Augmentation & None \\
Coordinate Range & $[-256,512]$ \\
Visibility Mask & Enabled \\
\bottomrule
\end{tabular}
\end{table}

\section{Baseline Model Details}
\label{appendix:full_pretraining_details_appendix}

\subsection{VideoMAEv2 Pre-training Details}
We use the official VideoMAEv2-Base model checkpoint pre-trained for 800 epochs in a self-supervised way on UnlabeldHybrid-1M dataset. The pre-training dataset details can be seen in Table \ref{tab:mae_pretrain}.
\begin{table}[h]
\centering
\caption{\textbf{VideoMAEv2 Pre-training Dataset Overview.} \\}
\label{tab:mae_pretrain}
\begin{tabular}{@{} l r l @{}}
\toprule
Dataset & Size & Source \\
\midrule
K710 \citep{li2022uniformerv2} & 658k & YouTube \\
SSv2 \citep{ssv2} & 169k & Shot from Scripts \\
AVA \citep{gu2018avavideodatasetspatiotemporally} & 21k & Movie \\
WebVid2M \citep{Bain_2021_ICCV} & 250k & Internet \\
self-collected & 250k & Instagram \\
UnlabeledHybrid & 1.348M & Multi-Source \\
\bottomrule
\end{tabular}
\end{table}
\subsection{V-JEPA 2 Pre-training Details}
\label{appendix:vjepa_caution_note}
We use the official pre-trained V-JEPA 2 ViT-L model checkpoint. This is the smallest publicly available V-JEPA 2 checkpoint. As a result, the discrepancy between model sizes (i.e., ~80M parameters for other baselines vs ~300M for V-JEPA 2) may provide a competitive performance advantage to the V-JEPA 2 model during evaluation. The pre-training dataset details can be seen in Table \ref{tab:vm22m}.

\begin{table}[h]
\centering
\label{tab:vjepapretraining}
\caption{\textbf{\vjepa~VideoMix22M (VM22M) Pretraining Dataset.} \\}
\label{tab:vm22m}
\begin{tabular}{@{} l c c c c c @{}}
\toprule
Source & Samples & Type & Total Hours & Apply Curation & Weight \\
\midrule
SSv2 \citep{ssv2} & 168K & EgoVideo & 168 & No & 0.056 \\
Kinetics \citep{carreira2022shortnotekinetics700human} & 733K & ExoVideo & 614 & No & 0.188 \\
Howto100M \citep{miech2019howto100mlearningtextvideoembedding} & 1.1M & ExoVideo & 134K & No & 0.318 \\
YT-Temporal-1B \citep{zellers2022merlotreserve} & 19M & ExoVideo & 1.6M & Yes & 0.188 \\
ImageNet \citep{deng2009imagenet} & 1M & Images & n/a & No & 0.250 \\
\bottomrule
\end{tabular}
\end{table}

\subsection{RVM Pre-training Details}

We use the official RVM ViT-B pre-trained checkpoint from Google DeepMind public repository. The pre-training dataset details can be seen in Table \ref{tab:rvm_pretrain}.

\begin{table}[h]
\centering
\caption{\textbf{RVM Pretraining Dataset Composition.} \\}
\label{tab:rvm_pretrain}
\begin{tabular}{@{} l c c c c c @{}}
\toprule
Source & Samples & Type & FPS & Apply Curation & Weight \\
\midrule
SSv2 \citep{ssv2} & 168K & EgoVideo & 25 & No & 0.056 \\
Kinetics-700 \citep{carreira2022shortnotekinetics700human} & 733K & ExoVideo & 25 & No & 0.188 \\
Howto100M \citep{miech2019howto100mlearningtextvideoembedding} & 1.1M & ExoVideo & 10 & No & 0.318 \\
YT8M \citep{abuelhaija2016youtube8mlargescalevideoclassification} & 3.3M & ExoVideo & 10 & No & 0.188 \\
YT-BoundingBoxes \citep{real2017youtubeboundingboxeslargehighprecisionhumanannotated} & 380K & ExoVideo & 10 & No & 0.250 \\
\bottomrule
\end{tabular}
\end{table}

\subsection{Linear Probe Baseline Feature Dimensions}

\label{appendix:linear_probe_feature_dimensions}

To ensure a fair comparison between model features, we maintain the standard linear probing protocol in self-supervised learning - mean pooling over the token sequence. We report the final resulting feature vector dimensions in Table~\ref{tab:linear_probe_feature_dimensions_table}.

While all the video models (including TIME) used in our evaluation are base transformer size-equivalent (ViT-B), the smallest publicly available V-JEPA 2 checkpoint is ViT-L. This discrepancy between model sizes (i.e., ~80M parameters for other baselines vs ~300M for V-JEPA 2) may provide a competitive performance advantage to the V-JEPA 2 model during evaluation.

\begin{table}[h]
\centering
\caption{\textbf{Feature readout protocol.} All models are compared using standard linear probing with mean pooling over the token sequence. \\}
\label{tab:linear_probe_feature_dimensions_table}
\begin{tabular}{@{}llc@{}}
\toprule
\textbf{Model} & \textbf{Architecture} & \textbf{Mean-pooled Feature Vector Dimension} \\
\midrule
TIME & ViT-B & 768 \\
VideoMAEv2 & ViT-B & 768 \\
RVM & ViT-B & 768 \\
V-JEPA 2 & ViT-L & 1024 \\
\bottomrule
\end{tabular}
\end{table}

\section{Evaluation Design Choices}
\label{appendix:further_evaluation_decisions}

\subsection{SSv2 ``Arrow of Time'' class pairs}
\label{appendix:aot_pairs}

We provide the action pair details for the 28 classes of the Something-Something v2 \citep{ssv2} ``Arrow of Time'' subset in Table \ref{tab:action_pairs}.

\begin{table}[h]
    \centering
    \caption{\textbf{SSv2 ``Arrow of Time'' Forward and Backward Action Pairs.\\}\\}
    \label{tab:action_pairs}
    \begin{tabular}{ll}
        \toprule
        \textbf{Forward Action} & \textbf{Backward Action} \\
        \midrule
        Moving something up & Moving something down \\
        Pushing from left to right & Pushing from right to left \\
        Pulling from left to right & Pulling from right to left \\
        Moving towards the camera & Moving away from the camera \\
        Moving things closer & Moving things away \\
        Moving closer to something & Moving away from something \\
        Lifting up completely & Putting down completely \\
        Lifting up one end & Putting down one end \\
        Rolling down a slant & Rolling up a slant \\
        Opening something & Closing something \\
        Folding something & Unfolding something \\
        Covering something & Uncovering something \\
        Putting into something & Taking out of something \\
        Plugging into something & Pulling out of something \\
        \bottomrule
    \end{tabular}
\end{table}

\subsection{Ego-Exo4D Preprocessing}
\label{appendix:egoexo_preprocessing}
For Ego-Exo4D \citep{grauman2024egoexo4d}, we use all available video takes containing \emph{keystep annotations} (i.e., ``Bike Repair'', ``Cooking'', and ``Health'' scenes), resulting in a set containing 24,060 training and 6,014 validation samples across 620 classes. We extract all action snippets from these Ego-Exo4D scenes containing the `best Exo' camera view as labeled in the metadata. We include all actions between 0.5s and 5.0s in length. If an action snippet is longer than 5.0s, we center crop these action snippets temporally to capture the most representative portion of the video.

\section{\emb~Model Efficiency Claims}
\label{appendix:model_efficiency_clarification}
\subsection{Training Efficiency}
We further explain the model efficiency by comparing to the gains of other video models when trained with more data. For example, as reported by the authors, V-JEPA trained on 2M videos vs trained 22M videos gains $0.7\%$ on average for temporal datasets (SSV2, Jester, Diving48) \citep{vjepa2}. In other words, it takes 20M videos to increase $0.7\%$. On the datasets that we experimented with TIME (SSV2, Diving48, EgoExo4D – note we exclude Jester as accuracies around $97\%$ are practically saturated), we are able to improve V-JEPA’s performance by up to $4.15\%$ using 250K synthetic videos. This points to the fact that TIME is able to extract information in an alternative and complementary manner, which is more efficient than increasing the amount of training data of a specific model.

\subsection{Addressing Point Tracking Penalty}
Use of point tracking models (i.e., CoTracker3~\citep{cotracker3}) involves additional training data used to train the models themselves. However, when put in context numerically, even this additional training is remarkably small compared to existing foundation models. TIME is trained on 250K videos, of 2 seconds each, and CoTracker3 teacher models are primarily trained on 10K videos of 5 seconds each. This is a total of 0.017 years of video, compared to V-JEPA 2, trained on over 100 years of video.

\section{Limitations}

In this paper we have shown the surprising power of training on motion information to learn video representations efficiently. We consider this work a stepping stone for several lines of future work that include the following. First, this model focuses on temporal information alone, and the integrations we have done with other appearance-based models are straightforward (e.g., in the form of concatenation). Fully integrating the \emb~representation into an appearance-based model, while maintaining the temporal separation that has proved so advantageous  will be a promising area of research. Second, this model at inference time relies on point-tracking techniques. While recent research~\citep{cotracker3} has made them faster and more accurate, they might still introduce errors in fast-moving or blurry regions. Integrating this embedding with a point-tracking system or using other motion representations will also be a very promising direction.

\section{Synthetic Data Visualizations}
\label{appendix:additional_viz_appendix}
\begin{figure}[H]
  \centering
  \begin{minipage}[b]{1.0\textwidth}
    \includegraphics[width=\textwidth]{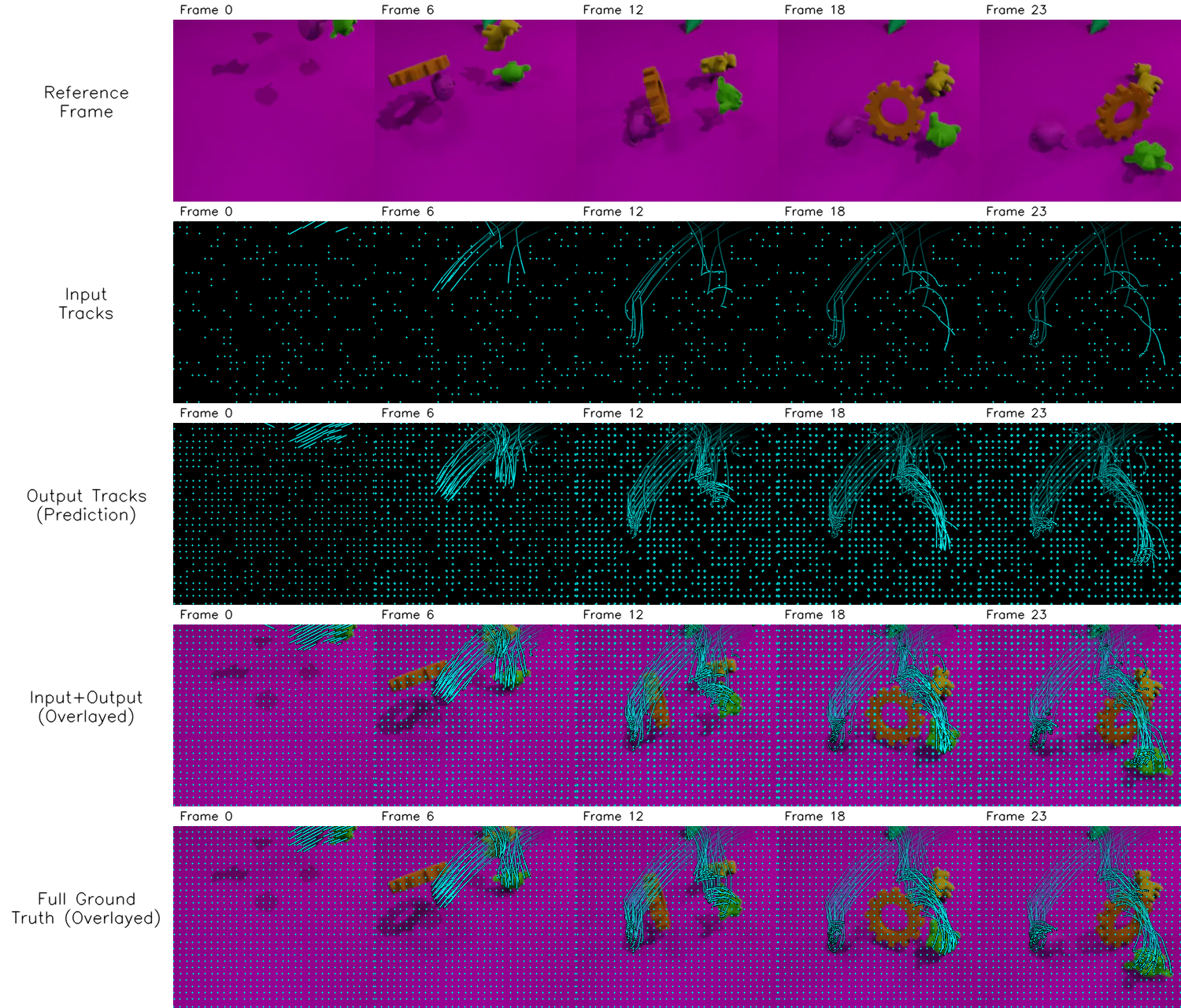}
  \caption{TIME visualization. Our model reconstructs missing trajectories which closely align with the ground truth.}
  \label{fig:additional_viz}
  \end{minipage}
\end{figure}

\begin{figure}[H]
  \centering
  \begin{minipage}[b]{1.0\textwidth}
    \includegraphics[width=\textwidth]{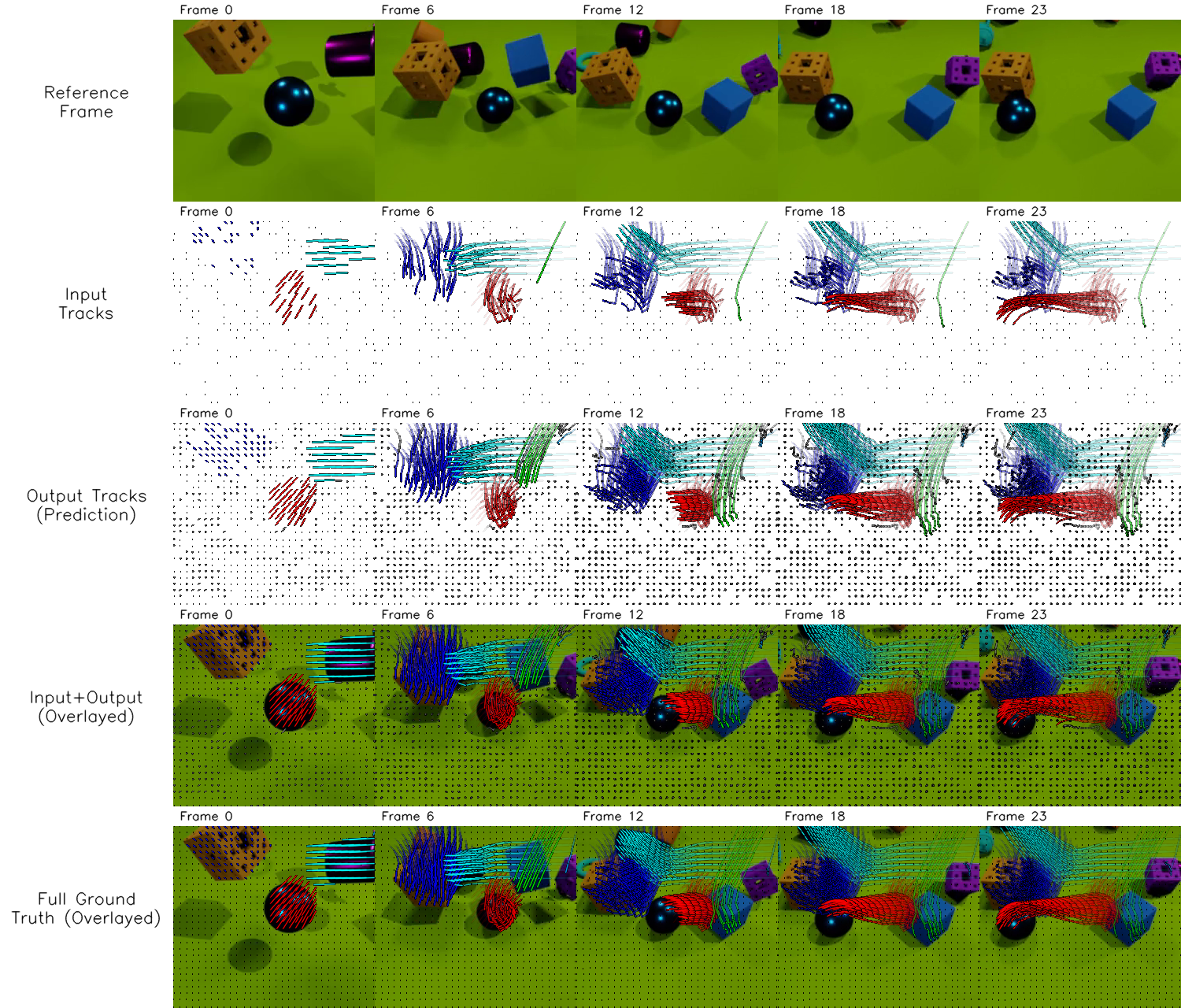}
    \caption{TIME visualization (objects segmented by colors). In this visualization we can observe that the proposed model successfully captures multi-object collisions and velocity changes. Static point tracks that fall onto the background remain static throughout the timeline.}
    \label{fig:additional_viz_segmented}
  \end{minipage}
\end{figure}

\begin{figure}[H]
  \centering
  \begin{minipage}[b]{1.0\textwidth}
    \includegraphics[width=\textwidth]{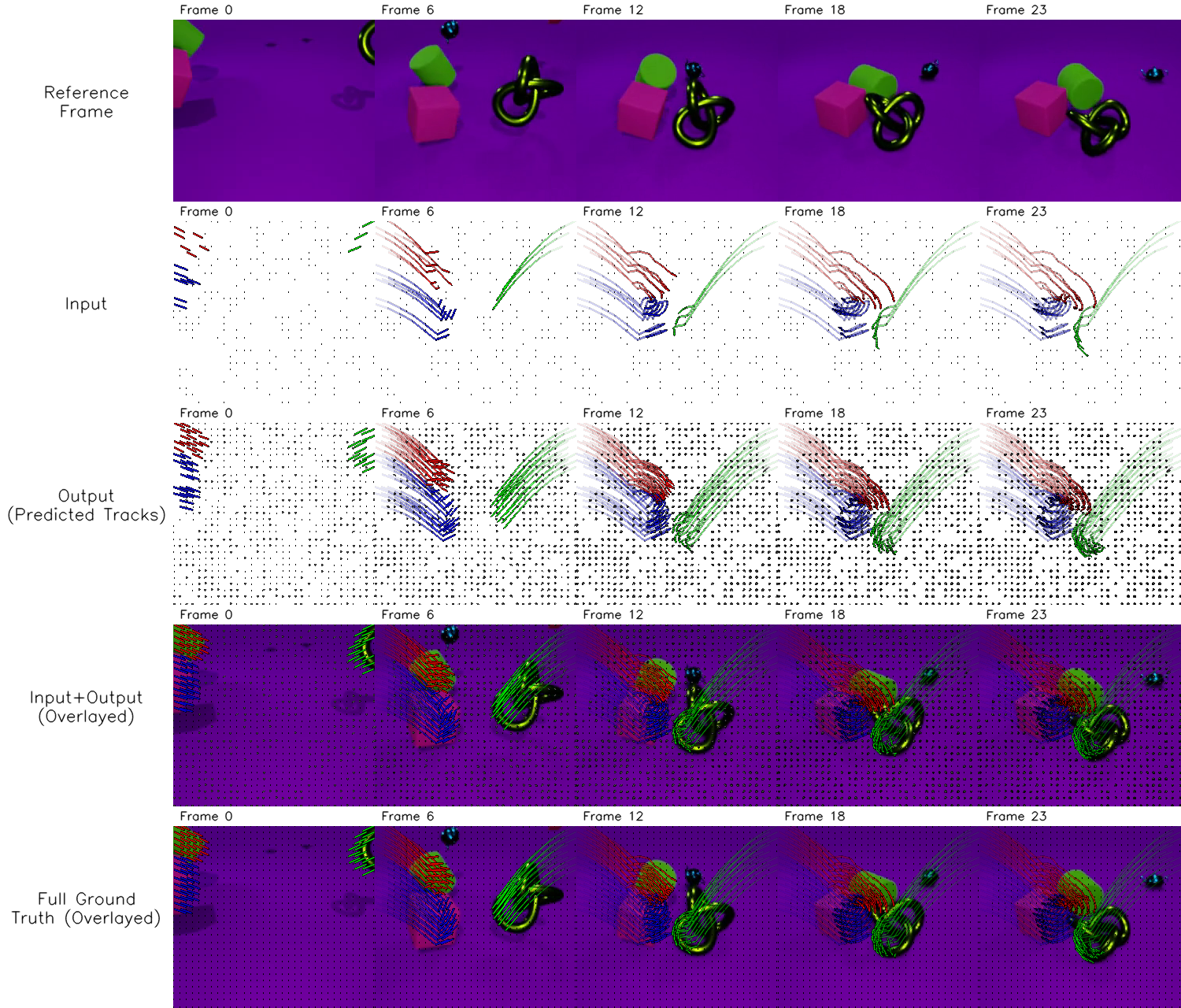}
  \caption{TIME visualization (objects segmented by colors). In this visualization we can observe that the proposed model successfully captures angular rotation of a complex object (e.g., torus knot).}
  \label{fig:additional_viz_segmented2}
  \end{minipage}
\end{figure}

\begin{figure}[H]
  \centering
  \begin{minipage}[b]{1.0\textwidth}
    \includegraphics[width=\textwidth]{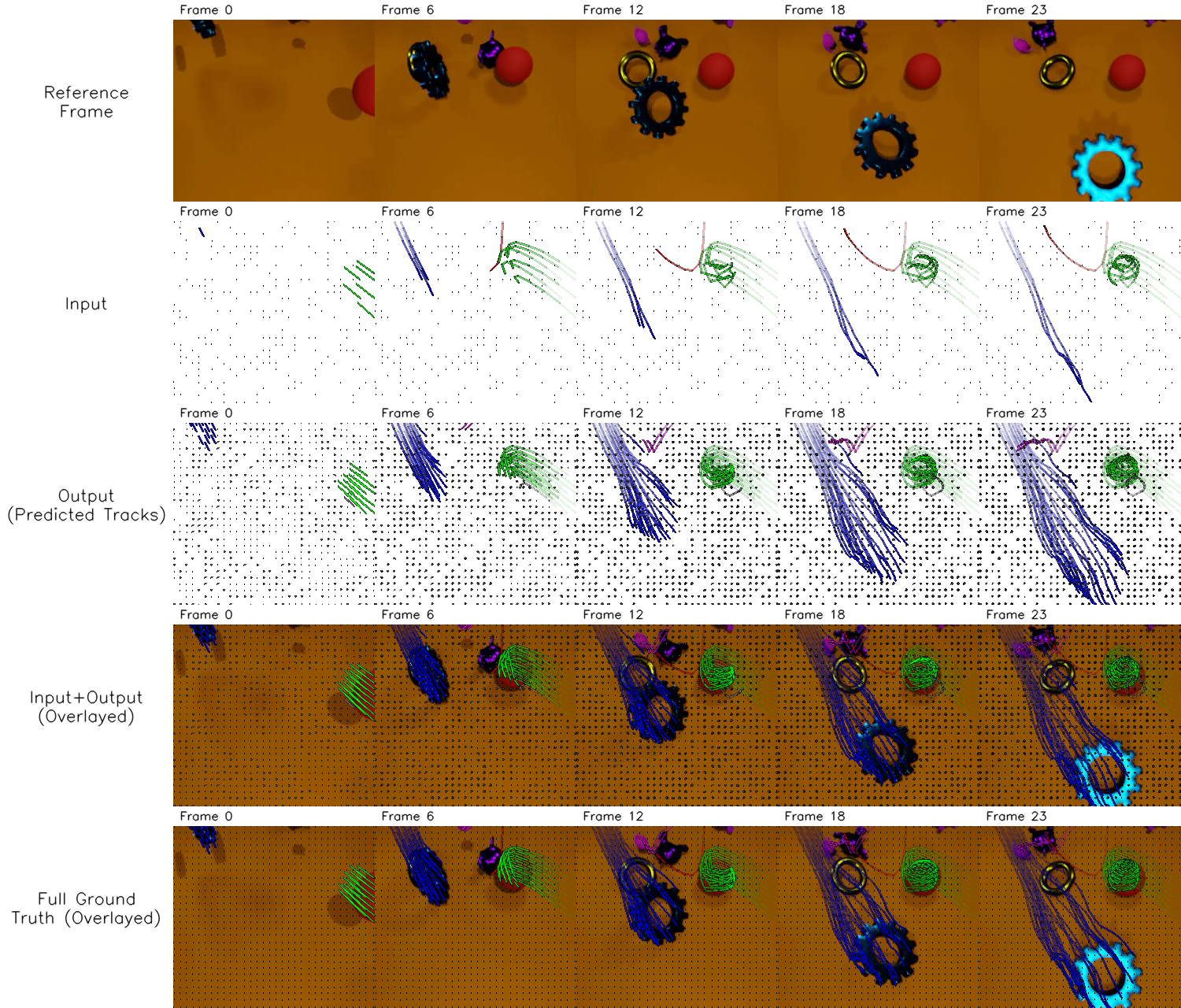}
    \caption{TIME visualization (objects segmented by colors). The proposed model successfully reconstructs object scale, rotation and velocity changes.}
    \label{fig:additional_viz_segmented3}
  \end{minipage}
\end{figure}

During the experiments we observe that point tracks can capture events such as object collisions, interactions, velocity, scale changes, rotation and acceleration. We include a more extensive visualization of our model trajectory reconstruction progression in Figure \ref{fig:additional_viz}. We also provide similar visualizations and briefly describe observed motion principles where objects are also segmented by colors (Figures \ref{fig:additional_viz_segmented},  \ref{fig:additional_viz_segmented2},  and  \ref{fig:additional_viz_segmented3}).

\clearpage 
 
\clearpage 

\clearpage 

\clearpage 

\end{document}

%% file: math_commands.tex
\usepackage{amsmath,amsfonts,bm}

\def\eqref#1{equation~\ref{#1}}

\def\1{\bm{1}}

\DeclareMathAlphabet{\mathsfit}{\encodingdefault}{\sfdefault}{m}{sl}
\SetMathAlphabet{\mathsfit}{bold}{\encodingdefault}{\sfdefault}{bx}{n}



%% file: abstract.tex
\begin{abstract}
Video representation learning has seen tremendous progress in recent years. This has been driven by many factors, including the scale of training and the success of self-supervised models trained on next-frame prediction. While these factors have pushed the boundaries of what video models can do, they also introduce their own set of limitations. First, scaling video models can reach prohibitive costs, with recent models needing hundreds of years of video to be trained. Second, learning to predict the next frame (or embedding of the frame) naturally focuses on spatial information and as a result, video models still struggle with temporal understanding. 
In this paper we propose a novel approach that uses motion as a modality to alleviate both of these core issues. 
Given the motion in a video in the form of point-tracks, we mask some of the tracks and use a masked autoencoder to reconstruct the missing tracks. This allows us to learn a representation in a self-supervised manner which we call {\emph \emb~}(Temporally-Informed Motion Embedding), and is focused on capturing temporal information. Also, as motion is inherently appearance-invariant, \emb~needs far fewer examples to generalize well. 
As a result, without bells and whistles, on temporal tasks \emb~performs on par with state-of-the-art models, using up to 4 orders of magnitude less training data. For general tasks, \emb~can be used in combination with existing representations, and we observe that it leads to a significant improvement for \vjepa, RVM and VideoMAE on standard benchmarks such as SSV2, EgoExo4D and Diving48. These results point to a new promising video paradigm for both more temporally-aware as well as more scalable models.



\end{abstract}

%% file: introduction.tex
\section{Introduction}

Video Understanding has seen great progress in the last few years. Some tasks, such as action recognition (e.g., Kinetics~\citep{kinetics}) or general question-answering (e.g., NExT-QA~\citep{xiao2021next}) 
can be performed extraordinarily well by current video models. 

This impressive progress has been possible thanks to several factors. One of the key factors is clearly the scale of training data. Similar to other areas in vision, the transition from pre-training on a single large dataset (e.g., Sports1M~\citep{sports1m}) to pre-training on vast amounts of data~\citep{vjepa2} has brought the capabilities of video models to a new level. However, scaling video training presents its own limitations in the trade-off between cost and accuracy. For example, recent examples of models~\citep{vjepa2} show that increasing the size of the training data by millions of videos may only improve accuracy by 1-2\%, pointing to a possible performance plateau ~\citep{bordes2025intphys2benchmarkingintuitive} despite the use of larger and more diverse datasets and scaling laws. 

Another key factor for progress in recent years has been the success of models trained on next-frame or next-frame-embedding prediction~\citep{vjepa2,recurrentvideo}. These self-supervised models avoid the limitation of using language as a supervision signal~\citep{clip}, which may limit the range of learned concepts to those appearing in the captions used for training. In these self-supervised models spatial accuracy is explicitly rewarded, but temporal accuracy is only learned as a byproduct. As a consequence, representing temporal dynamics and especially detailed fine-grained motion information may pose a challenge for these models.


\begin{figure}[!tbp]
  \centering
  \begin{minipage}[b]{0.49\textwidth}
    \centering
    \includegraphics[width=0.9\textwidth]{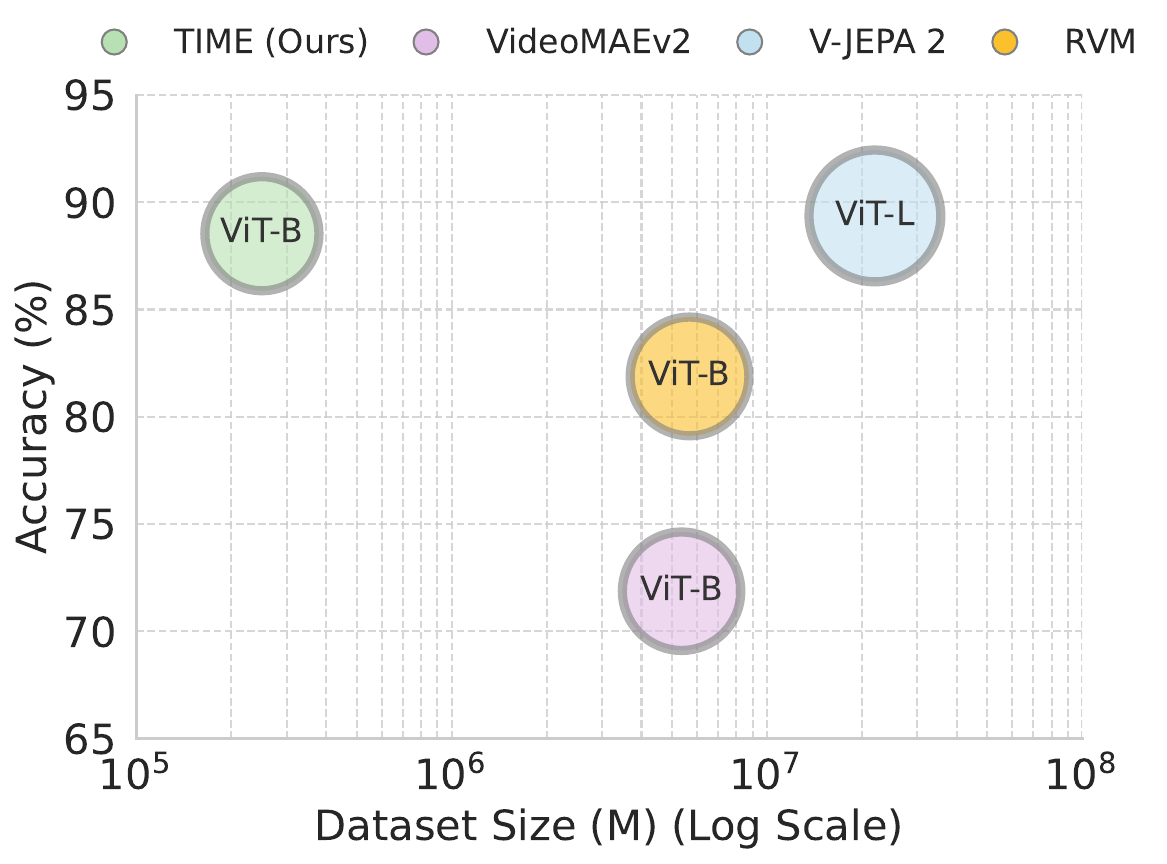}
    \caption{Model performance on SSv2 ``Arrow of Time'' task performance. Our TIME model achieves ``appearance-free'' action classification performance on-par with state-of-the-art \vjepa~model despite using several magnitudes less pre-training data.}
  \end{minipage}
  \hfill
    \begin{minipage}[b]{0.49\textwidth}
    \centering
    \includegraphics[width=0.88\textwidth]{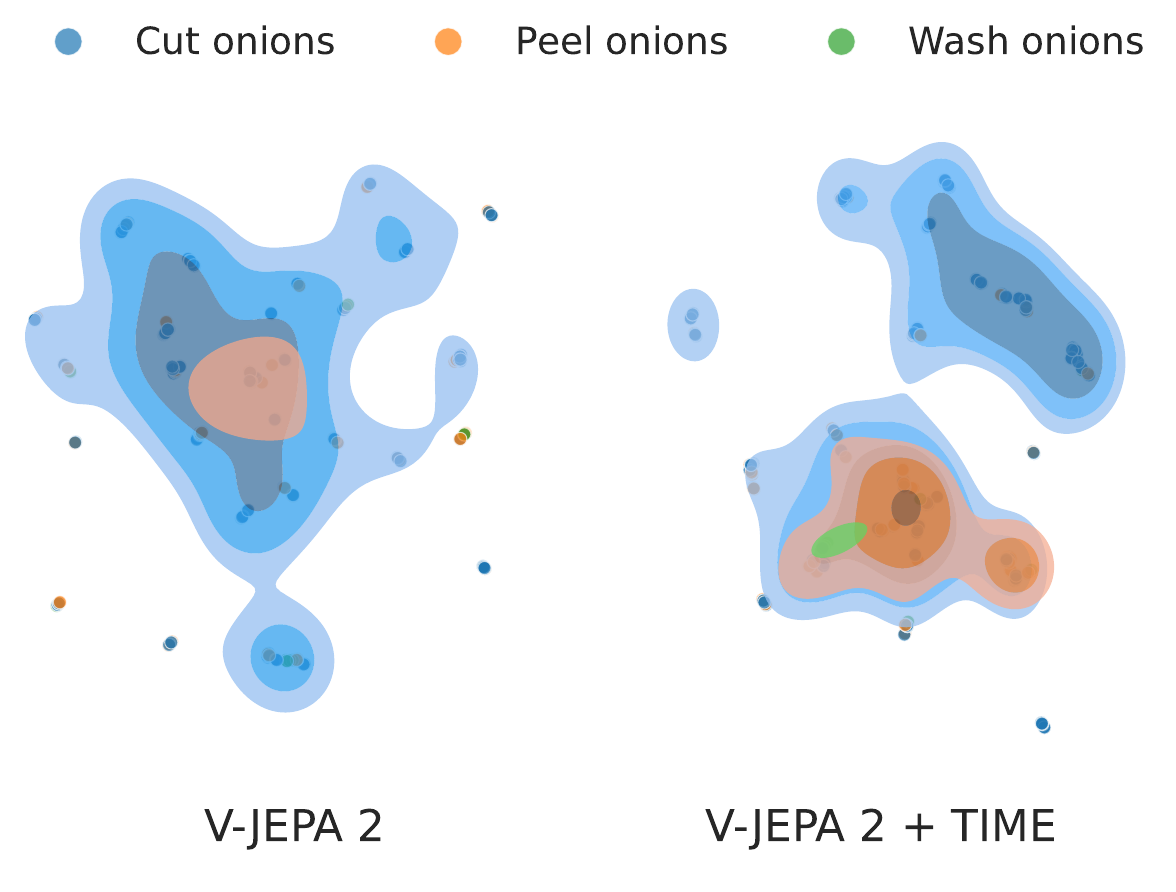}
      \caption{UMAP plot ($n\_components=2$) showing class separation for \vjepa~vs. concatenating  \vjepa~and \emb, on frozen features of videos from Ego-Exo4D. Using \emb~greatly improves the separation of similar fine-grained actions.}
  \end{minipage}
\end{figure}

The struggle to capture temporal dynamics has also been observed in Multi-modal Large Language Models (MLLMs), with a variety of new benchmarks illustrating this problem (MotionBench~\citep{motionbench}, TempCompass~\citep{tempcompass}, VideoMME~\citep{videomme}, VLM4D~\citep{vlm4d}). 

In this paper we make the observation that these two issues in video can be addressed with a key tool: learning video representations from motion. In particular, we use the motion of a video, in the form of point-tracks (sparsely sampled) as the only input modality to a masked-autoencoder. The input tracks are masked, and the training objective is to reconstruct the missing tracks. Crucially, the model is trained exclusively on synthetic point trajectories from rigid-body physics simulations from Kubric~\citep{kubric}, and has not seen any real-world video. The learned representation, which we call \emb~{\em(Temporally Informed Motion Embedding)}, can then be used for a variety of tasks in a zero-shot manner. In addition, it can be used as a complementary modality for appearance-based video models (e.g., VideoMAE~\citep{videomae}, \vjepa~\citep{vjepa2}) showing that the information that \emb~captures is complementary, and non-overlapping with these appearance-based models. The benefits of \emb~are twofold.  

First, we experimentally observe that \emb~improves performance on standard benchmarks (e.g., SSv2, Ego-Exo4D) when added to \sota~models such as \vjepa~\citep{vjepa2} and RVM~\citep{recurrentvideo}. This is particularly striking in fine-grained tasks within Ego-Exo4D, where the use of \emb~improves performance up to 18\%. This confirms that fine-grained temporal information is hard to capture by appearance-based models. We also test the performance of ~\emb on its own. Since there are no established benchmarks that focus solely on temporal information, we test on a split of two existing benchmarks: CLEVRER~\citep{clevrer} and Something-something V2 (SSv2)~\citep{ssv2}, as it is often the case for temporal classes~\citep{bagad2025chiralityactiontimeawarevideo}. For SSv2 we test on classes that can be discriminated without image information (e.g., opening vs. closing). For CLEVRER we also test on tasks that do not require temporal information, such as collision timing. The splits and tasks will be available upon publication. Across tasks, \emb~matches or outperforms \sota~models such as \vjepa~and RVM.





Second and crucially, learning from motion reduces the amount of data needed to train the model by up to 4 orders of magnitude (measured in hours of video training data), compared to the same \sota~appearance-based video models. This reduction stems from several facts: 1) Motion can be learned from synthetic data as the domain gap between natural and synthetic motion is surprisingly small~\citep{flownet}. In other words, noise-less synthetic data allow models to converge faster than noisy real-world data. 2) Low-dimensional motion is more discriminative than low-dimensional images~\citep{davison2026, whathappensnext}. This implies that motion can be represented as a sparser set of points (e.g., a $32\times32$ point trajectory grid) instead of a high-resolution equivalent (e.g., $256\times256$ pixel image) enabling more efficient training. 3) Learning strong temporal information within an appearance-based model is hard, as most training tasks can be solved relying heavily on appearance information, even when trained on extremely large amounts of data. Hence, separating the learning process of temporal information guarantees that the model cannot ``cheat” by looking at pixel values or other appearance representations.

In summary, the proposed \emb~video representation provides a novel, highly scalable and compute-efficient paradigm for future research in video understanding.


%% file: related_work.tex
\section{Related Work}

\paragraph{Self-Supervised Video Models.}

The first video models trained with large-scale datasets~\citep{sports1m,kinetics} used classification supervision for training. While these models~\citep{kinetics,timesformer} have worked well, they require large amounts of labeled training data, which is expensive to produce. In addition, the rise of ``foundation models'' in other domains such as language or image~\citep{dino}, that are designed to be used in a ``zero-shot'' fashion, without the need to fine-tune, has created increasing interest in self-supervised video models. Research in this area has led to a wide variety of pre-text tasks. Some works~\citep{Recasens_2021_ICCV, Ranasinghe_2022_CVPR} have explored the use of spatial and temporal transformations, in combination with a contrastive learning paradigm where a video and a transformed copy should be represented as similar as possible. While this works well, the contrastive paradigm can be sensitive to the choice of negative pairs. In the search of more efficient video model pre-training, the notion of masked autoencoders (MAE) \citep{he2021maskedautoencodersscalablevision} in image domain have been adopted for self-supervised video model pre-training tasks as well. More precisely, works such as VideoMAE \citep{videomae, wang2023videomaev2scalingvideo} propose models that learn effective spatiotemporal representations by reconstructing masked portions of raw frame spatio-temporal patch tubes as their primary training objective. While such models achieve state-of-the-art results in some video understanding tasks, the very nature of such data modelling inherently biases the network to memorize high-frequency spatial details (textures, colors) rather than abstract motion. To address this issue, research works on learning efficient video representations include joint-embedding predictive architectures (JEPA) that forecast missing latent spatiotemporal regions \citep{vjepa2}. Authors reason that such a model is no longer forced to to reconstruct exact pixels, therefore it is forced to learn high-level semantic and kinematic abstractions instead. However, despite pre-training on 114 years-equivalent of curated video data, such models severely underperform in physical understanding tasks \citep{bordes2025intphys2benchmarkingintuitive} with evaluation performance being close to a random chance. While latest self-supervised learning works introduce an explicit sparse motion signal during full model pre-training as an effective way to encode temporal dynamics in the learned representations \citep{trackMAE}, the sparsity of the signal and the requirement to retrain the entire model from scratch hinders its applications further.

\paragraph{Challenges in Temporal Understanding.}

Learning temporal information has been a long-standing challenge for video research. Historically, video models have tended to have stronger image understanding than temporal understanding capabilities. There are potentially several reasons for this, one of which is the challenge of designing tasks that force models to learn temporal information, without resorting to image information alone~\citep{onlytime, bagad2025chiralityactiontimeawarevideo}. Even designing benchmarks that allow us to measure temporal capabilities alone is not trivial. Several datasets~\citep{ssv2, diving48, SkatingCoach} have tried to focus on actions where appearance carries limited information. Others have explored tasks that are action independent such as counting~\citep{countingovr}, skill determination~\citep{skilldetermination}, adverb prediction \citep{doughty2020action,Moltisanti_2023_CVPR}, for tasks that are fairly object-independent such as the reasoning benchmark CLEVRER~\citep{clevrer}. Overall, solving these remains an open problem, as for example the accuracy of top models on the rather ``temporal'' Something-something dataset published in 2017, still only achieves values around 75\%, while accuracy on the rather ``appearance-based'' Kinetics published the same year achieves around 90\%.

In the new era of Multi-Modal Language Models (MLLMs), new  challenges in temporal understanding have arisen. The key obstacles might be different now, and relate to the reliance of language in the learning objective, or to token dilution. Still, it has been noted that MLLMs notoriously struggle with understanding temporal concepts~\citep{timeblindness, xue2025seeingarrowtimelarge}, and there have been a large variety of multi-modal benchmarks (TempCompass~\citep{tempcompass}, MotionBench~\citep{Hong_2025_CVPR}, TemporalBench~\citep{temporalbench}, VideoMME~\citep{videomme}, VLM4D~\citep{videomme}). All these efforts point out that temporal understanding is very much a one of the main challenges in video research today, across different technology paradigms. 

\paragraph{Learning Motion Models from Synthetic Data. }

While learning from synthetic data might seem like a limiting choice due to the sim-to-real domain gap, such limitation is significantly less restrictive for temporal kinematics than it is for spatial appearance. In fact, there is a long history of learning motion from synthetic data that shows surprisingly good results. One of the first success stories was FlowNet~\citep{flownet}, where an optical flow model was learned from the FlyingChairs dataset, consisting of 3D models of chairs moving around with an image background. Despite its simplicity or lack of realism, training on this dataset was a breakthrough on what could be learned from purely synthetic data, which led to other work leveraging this insight~\citep{autoflow}. In a different domain, point-tracking systems have also leveraged this. For example, the state-of-the-art dense point tracker model CoTracker~\citep{cotracker3} is also primarily trained using synthetic data from a simulated environment~\citep{kubric} without loss of real-world generality. Work in generative video modeling argue that current video model inability to model physics correctly arises from the overhead of generating pixels instead of motion, therefore, they rely on the very same synthetic Kubric environments and point tracking modules to obtain quasi-dense trajectories for the task of forecasting the future trajectories \citep{whathappensnext} from a single input image. Remarkably, coupled with a single-frame spatial context signal from DINO \citep{dino}, such models surpass expensive generative video models.

However, these prior methods exclusively use synthetic data to solve low-level vision tasks (e.g., pixel displacement or point tracking). In this work, we go one step beyond to demonstrate that purely synthetic simulations can be used to learn high-level semantic video representations that are generalizable and directly transferable to real-world scenarios.  While we initially presumed a real-world fine-tuning stage would be necessary to bridge the gap to complex human actions, surprisingly, our synthetic kinematic model transfers entirely zero-shot to real-world videos (e.g., SSv2~\citep{ssv2}). Operating out-of-the-box, it not only matches, but frequently outperforms and complements massively scaled, foundation video models. 

%% file: methods.tex
\section{Methods}

\begin{figure}[t]
    \centering
    \includegraphics[width=0.88\linewidth]{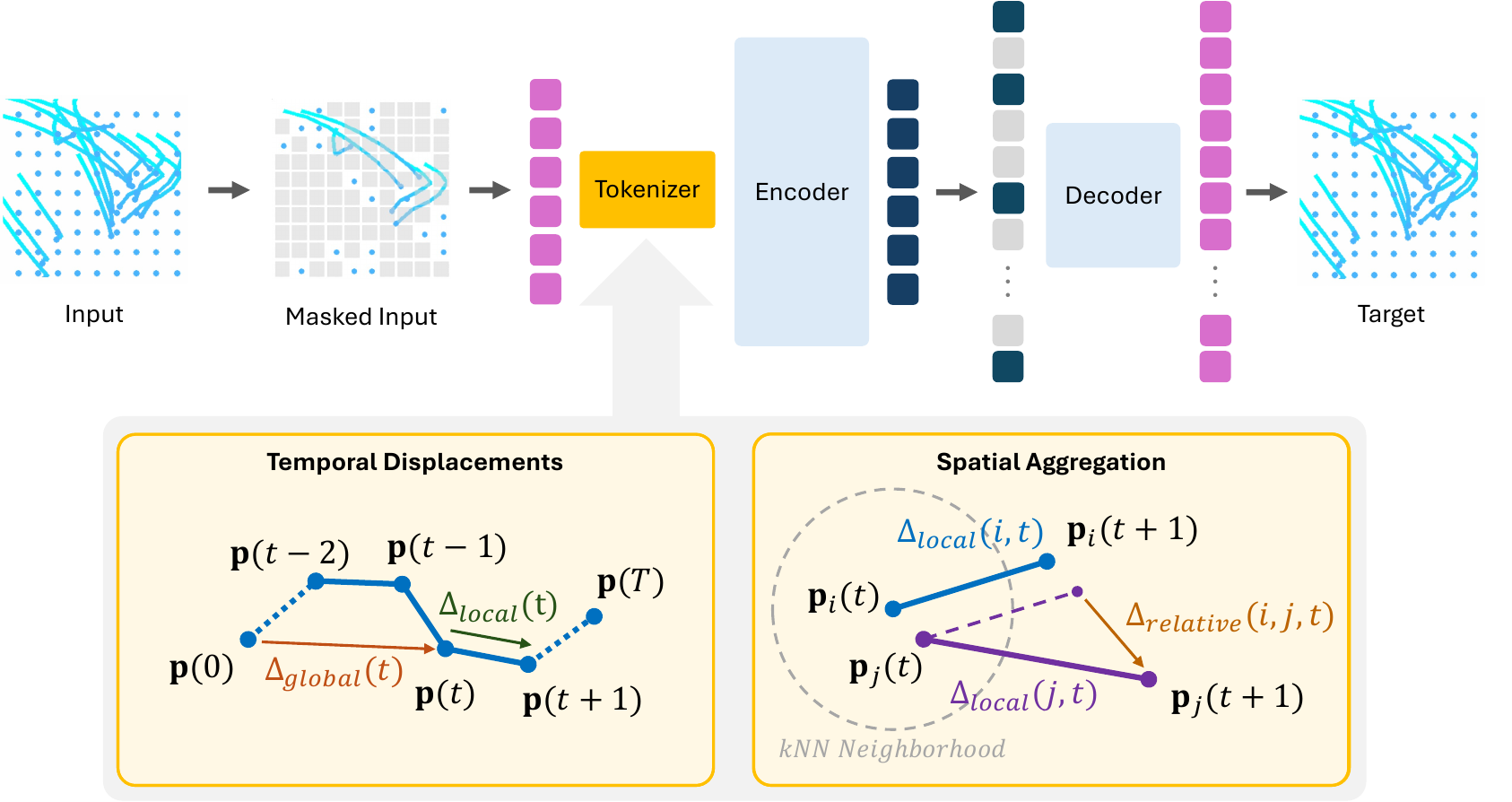}
    \caption{\emb~Architecture. Given a set of point trajectories, the model groups them into tubelets and encodes them into trajectory tokens using a tokenizer. During training, we mask 75\% of the tokens and only the visible ones are passed through the encoder. A decoder then attempts to reconstruct the masked trajectory tokens from the encoder output. The training loss is computed only on the masked trajectories.}
    \label{fig:time_architecture}
\end{figure}

\paragraph{Architecture Overview.} We now describe the details of the proposed \emb~architecture, which is illustrated in Fig.~\ref{fig:time_architecture}. Recall that the goal of this architecture is to learn a video representation that will capture the temporal content of the video. For this, we use a self-supervised training paradigm, which allows us to bypass the restrictions of learning from video captions. In particular, given the point tracks of a scene, sampled sparsely but in a grid, we mask a large set of them spatially (i.e., we mask the full point track, not just a portion or segment in time). We convert these point tracks to tokens, using a tokenization process to capture both spatial context of the tracks as well as local and global temporal information (details described below). These tokens are input to a masked autoencoder, similar to previous work such as VideoMAE, composed by an encoder and decoder. This autoencoder is trained using the full set point tracks as target for the reconstruction. At inference time, given a video, we compute the point tracks over the entire scene using point tracking methods~\citep{cotracker3}, and use them as input to the tokenizer, and in turn to the encoder. The resulting output from the encoder is a feature vector with strong capabilities to capture the temporal information of the scene. We now describe each of these components in detail.    

\paragraph{From Point Tracks to Tokens. } Our model operates over a grid of point tracks sampled uniformly. In our experiments, we use a grid of $32\times32$ point tracks, over a length of $24$ frames. Therefore the input to our model is $N = 1024$ points. Let $p_i(t) = (x,y)$ be the location in image coordinates of each of these points $p_i$ at time $t$. For every point $p_i$ we compute its corresponding token $s_i$ (for segment) to include essential information that may be useful for reconstruction. First, we compute the local displacement of the point to the next time step, $\Delta_\text{local}(i, t) = p_i(t+1) - p_i(t)$. Note this is essential, as using as input absolute point locations instead of relative displacements leads to collapse during training. Another failure mode could be for the network to accumulate small errors over a point track, in the end leading to a poor estimate. To prevent this, we incorporate information in the tokens about the global displacement $\Delta_\text{global}(i, t) = p_i(t) - p_i(0)$. This process is described in Fig.~\ref{fig:time_architecture} in the ``Temporal Displacements'' box. Capturing spatial structure and geometry is also helpful for point track reconstruction, for example for establishing motion boundaries. To incorporate this information, we take inspiration from the literature in point cloud self-supervised learning~\citep{pang2022maskedautoencoderspointcloud}. In particular, for a given point in time $p_i(t)$ and its given local displacement $\Delta_\text{local}(i, t)$, we compute how much the displacements of the spatial neighborhood around $p_i$ deviate from its local displacement. This process is described in Fig.~\ref{fig:time_architecture}, in the box ``Spatial Aggregation''. Specifically, given the $K$ nearest neighbors $p_j(t)$ of anchor point $p_i(t)$, we compute the relative displacement between each pair as $\Delta_{\text{relative}}(i,j,t) = p_j(t+1) - p_i(t+1)$. We do max-pooling over the relative displacements, to obtain the maximum deviation $\Delta_{\text{max-deviation}}(i,t)$, which will also be part of the token. In our experiments we use $K=16$. Finally, we include an occlusion bit $o_i(t)$, with information of whether point $p_i$ is occluded or not. In summary, for point $p_i(t)$, its corresponding token $s_i(t)$ is computed as: 
\begin{equation}
    s_i(t) = [\Delta_\text{local}(i, t), \, \Delta_\text{global}(i, t), \, \Delta_{\text{max-deviation}}(i,t),\, o_i(t)].
\end{equation}

\begin{figure}[t]
    \centering
\includegraphics[width=0.95\linewidth]{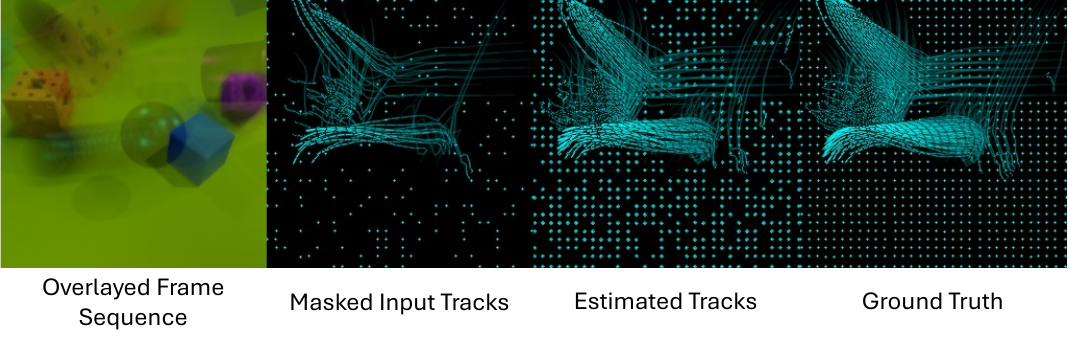}
    \caption{{\bf Sample Training Scene.} Given the input tracks of a scene created with Kubric~\citep{kubric}, the proposed architecture is able to fill in the gaps and estimates the masked tracks using a masked autoencoder, with high fidelity. } 
\label{fig:sample_scene}
\end{figure}



\paragraph{Factorized Spatiotemporal Attention Backbone.}
Following the tokenization steps, we feed the resulting tokens into a model largely based on VideoMAE~\citep{videomae}. Our encoder and decoder contains 12 and 4 layers respectively. In addition, we factorize the attention to improve computational efficiency. More specifically, as the original ViT relies on \textit{full spatio-temporal attention}, its time complexity is expensive at $O((T \cdot S)^2)$. We leverage successful applications of \textit{factorized attention} for point tracking \citep{cotracker3} and image-to-trajectory forecasting \citep{whathappensnext}, and factorize the attention mechanism as well. First, we compute \textit{spatial attention} across all unmasked trajectory tokens within the same pair of frames. Next, we compute temporal attention across the time dimension for each individual trajectory. The resulting separation of spatial and temporal components reduces attention time complexity to $O(T \cdot S^2 + S \cdot T^2)$ leading to a five-fold reduction in training time. Further tokenization details can be found in Appendix~\ref{appendix:tokenization_details}.

\paragraph{Training Objective.}
The proposed model is trained to reconstruct masked trajectory tokens from visible context (Fig. \ref{fig:time_architecture}). During training, we mask $75\%$ of the spatial trajectory tokens and apply this mask across the entire temporal dimension. The encoder processes only the visible tokens and produces a representation of the video. The decoder then inserts learnable mask tokens for the missing trajectory positions and reconstructs the full sequence.

\paragraph{Loss Function.} 
We employ the Huber loss $\mathcal{L}_{\text{Huber}}$ (Smooth $L_1$) over masked tokens. We compute the loss of local and global displacements separately as they naturally have very different numerical values, and need to be weighted differently. We use a hyperameter $\lambda_{\text{global}}$ to weight down the values of the global displacements ($\lambda = 0.5$). 
The resulting loss target ensures a balance between local high-frequency kinematics and full temporal length-wide global motion:
\begin{equation}
\mathcal{L}_{\text{target}}(i) = \mathcal{L}_{\text{Huber}} \Big( \hat{\Delta}_{\text{local}}(i), \Delta_{\text{local}}(i) \Big) + \lambda_{\text{global}} \, \mathcal{L}_{\text{Huber}} \Big( \hat{\Delta}_{\text{global}}(i), \Delta_{\text{global}}(i) \Big).
    \label{eq:loss_calculation}
\end{equation}
The natural distribution of displacements is also highly imbalanced. Since over 80\% of points in the training data are static, there is a risk that the model might predict everything as static. Hence, we also weigh non-static points with a higher weight $\omega$, computed as: 
\begin{equation}
    \omega{(i)} = \begin{cases}
        \gamma & \text{if } \|\Delta_{\text{global}}(i)\|_2 > \tau \\
        1.0 & \text{otherwise}
    \end{cases}
    \label{eq:motion_multiplier}
\end{equation}
where $\tau$ is a pre-defined static motion threshold (  $\tau=0.002$), and $\gamma = 7.0$ represents the motion boost hyperparameter. The final reconstruction loss is calculated as the weighted average across all masked trajectory tokens:
\begin{equation}
    \mathcal{L} = \frac{1}{\sum_{i} \omega^{(i)}} \sum_{i} \omega^{(i)} \, \mathcal{L}_{\text{target}}^{(i)}.
    \label{eq:main_loss}
\end{equation}
Additional loss function design details are available in Appendix~\ref{appendix:loss_clarification}.

\paragraph{Training Data.}
To train our model, we modify the Kubric simulation environment \citep{kubric} to obtain a uniform grid of point trajectories. This choice has several advantages. First, it generates perfectly accurate and clean data i.e., without the noise from the real-world tracks obtained using point tracking. This helps the training procedure making it more stable and faster to converge. Second, using generated data allows us to speed up the generation of point tracks, as instead of estimated they are directly computed. We produce 250K videos using MOVi-B dataset generation pipeline consisting of scenes with static camera and a random number of geometric primitives and simple shapes colliding in the middle of the scene. Further settings can be found in Appendix~\ref{appendix:movi_pipeline_settings}. 

\paragraph{Training Details.}

Our largest \emb~ model is trained on 250K synthetic samples for 100 epochs, using 3 warm-up epochs with an effective batch size of 40. We set the learning rate to 1.5e-4 and use the AdamW optimizer with cosine annealing learning rate scheduling technique with a weight decay of 0.05. Training using 4 NVIDIA A100 80GB GPUs takes around 120 hours ($\approx 5$ days). Full training details can be found in Appendix \ref{appendix:full_time_pretraining_details_appendix}.

%% file: experimental_results.tex
\section{Experimental Results}


We evaluate the proposed representation to answer several questions. First, we test whether in fact \emb~ is able to capture temporal information that is general enough to succeed in a zero-shot task.  To this end we re-purpose standard video datasets (SSv2~\citep{ssv2}, CLEVRER~\citep{clevrer}) to create tasks where appearance is not necessary, and motion carries all the information needed for the task. Section~\ref{subsec:temporal} describes the details of these experiments. 

Second, for general visual tasks, we explore how well the \emb~embedding complements other video models. For this we experiment on full standard benchmarks (such as SSv2~\citep{ssv2}, Ego-Exo4D~\citep{grauman2024egoexo4d}, Diving48~\citep{diving48}) using \sota~video models and combining \emb~features with them. Details are shown in Sec.~\ref{subsec:spatial}. 

Finally, in Sec.~\ref{subsec:ablation} we perform ablation studies to understand the behavior of \emb~ and what factors (such as dataset size, masking ratio, etc) affect results most. 

\subsection{Experimental Details}

\paragraph{Datasets.} We use a suite of standard, diverse datasets to evaluate \emb. For video understanding tasks, we evaluate on standard benchmarks that are generally temporally rich, including:
the Something-Something-V2 (SSv2)~\citep{ssv2} dataset, the Diving48~\citep{diving48} dataset and Ego-Exo4D~\citep{grauman2024egoexo4d}. Note that for Ego-Exo4D, we use all available video takes containing \emph{keystep annotations} (i.e., ``Bike Repair'', ``Cooking'', and ``Health'' scenes), resulting in a set containing 24,060 training and 6,014 validation samples across 620 classes (full details in Appendix \ref{appendix:egoexo_preprocessing}). We report average accuracy as well as the standard deviation of 5 separate runs. 

For temporal understanding tasks there are no standard datasets. We use a subset of SSV2 and a subset of CLEVRER~\citep{clevrer}, covering the tasks that require strictly temporal information. Further details on the split are given in Sec.~\ref{subsec:temp_tasks}.

\paragraph{Video Model Baselines.} As baseline comparisons we choose several video representations. We use VideoMAEv2~\citep{videomae} as the architecture that the proposed embedding is most closely related. This allows us to make comparisons across modalities for comparable architectures. We also use \vjepa~\citep{vjepa2} and RVM~\citep{recurrentvideo} as representatives of state-of-the-art in self-supervised video representations. All models are of ViT-B size-equivalent except for ~\vjepa~which is ViT-L (refer to Appendix ~\ref{appendix:vjepa_caution_note}). A complementary comparison against motion-only baselines is available in Appendix~\ref{appendix:motion_only_baseline_results_comparison}.

\subsection{\emb~ for Video Understanding}
\label{subsec:spatial}

\paragraph{Description of Video Understanding Tasks.} We now explore the use of the proposed embedding for general visual tasks, that require both spatial and temporal information. Our goal here is to test the ability of \emb~to provide additional information to standard, appearance-based, video models. 
To ensure a fair comparison between model features, we maintain the standard linear probing protocol in self-supervised learning - mean pooling over the token sequence.  The final resulting feature vector dimensions are available in Appendix~\ref{appendix:linear_probe_feature_dimensions}. 
Further experiments with alternative probing can be found in Appendix~\ref{appendix:attentive_probe}. 
Further evaluation design choices can also be found in Appendix ~\ref{appendix:further_design_decisions}.

\input{tables/combined_video_understanding}


\paragraph{Results. }

Table~\ref{tab:combined_video_understanding} shows the results. We observe that \emb~provides performance gains across datasets and models, showing that it is very much providing complementary information. These gains are particularly noticeable in the in Ego-Exo4D, where \emb~improves the performance of \vjepa~close to 11\%. This suggest that \emb~might be particularly useful for smaller, non-rigid motions. This is a particularly remarkable result because \emb~has been trained on orders of magnitude less data than the other models and because it was solely trained on synthetic data. This result suggests that training image representations and temporal representations independently may be an interesting solution to prevent models from overly relying on appearance information to solve the training tasks. Standalone model performance and additional dataset analyses are available in Appendix \ref{appendix:appendix_additional_results}. Additional model efficiency clarification can be found in Appendix \ref{appendix:model_efficiency_clarification}. 

\subsection{\emb~ for Temporal Reasoning}

\label{subsec:temporal}

\input{tables/temporal_tasks_table}

\paragraph{Description of Temporal Tasks.} 
\label{subsec:temp_tasks}Standard video understanding benchmarks often combine spatial and temporal information. Hence there are no existing benchmarks focusing strictly on temporal information (i.e., the solution to the task requires only temporal information). Here we split the section of SSv2 and CLEVRER that is strictly temporal, to allow us to directly compare the temporal capabilities of video models. 

In SSv2 there are many classes that require appearance information to be identified. Thus, we select 14 pairs of classes that have directional labels, such as ``Moving something up'' vs. ``Moving something down'', for a total of 28 classes (full details in Appendix~\ref{appendix:aot_pairs}). This subset consists of 43,583 videos from the training set and 5,859 videos from the validation set. 
The task is to classify each pair using frozen features and a linear classifier. We average the accuracy across the 14 binary classifiers. We refer to this task as ``Arrow of Time'' borrowing the name from previous work~\citep{arrowoftime}. Results are shown in Table~\ref{tab:arrow_of_time}, in the row ``SSv2 (AoT)''.

We also use the CLEVRER~\citep{clevrer} dataset. This dataset is also designed to focus on temporal reasoning, but its question-answering would potentially involve the use of a multi-modal model, which could complicate the process of strictly measuring temporal abilities. Instead, we use the meta-data present in the videos to design three tasks: Task 1, Classification (``Does a collision happen?''); Task 2, Detection (``When does the first collision happen?''); Task 3, Counting (``What is the number of collisions?''). Each task is evaluated on the full dataset containing 10,000 training and 5,000 validation videos. The task is similar to before, to use frozen features to train a linear classifier or regressor. Results are shown in Table~\ref{tab:arrow_of_time}. 

\paragraph{Results.} The experiments show a surprising finding. \emb~has been trained only on synthetic data and on orders of magnitude less data than \sota~models, while those models have been trained on real data, which includes for example the SSv2 dataset. Concretely, compared to the \vjepa~model which is the most expensive to train, the comparison is remarkable, 140 hours of simulations vs. 200 years of real-world video respectively. Yet, \emb~performs on par or even surpasses the other self-supervised video models in temporal understanding. For the detection task in particular, the proposed \emb~has a gap of over 15\% over \vjepa. This is an interesting finding that suggests that despite training scale, current video model training paradigms might overly rely on visual appearance cues, hindering them from learning temporal structure. Instead, \emb~is forced to learn a strong kinematic representation, which shows excellent generalization abilities, even to unseen tasks, and real-world data.

\subsection{Ablation Study of \emb}
\label{subsec:ablation}

\begin{figure}[!tbp]
  \centering
  \includegraphics[width=0.9\textwidth]{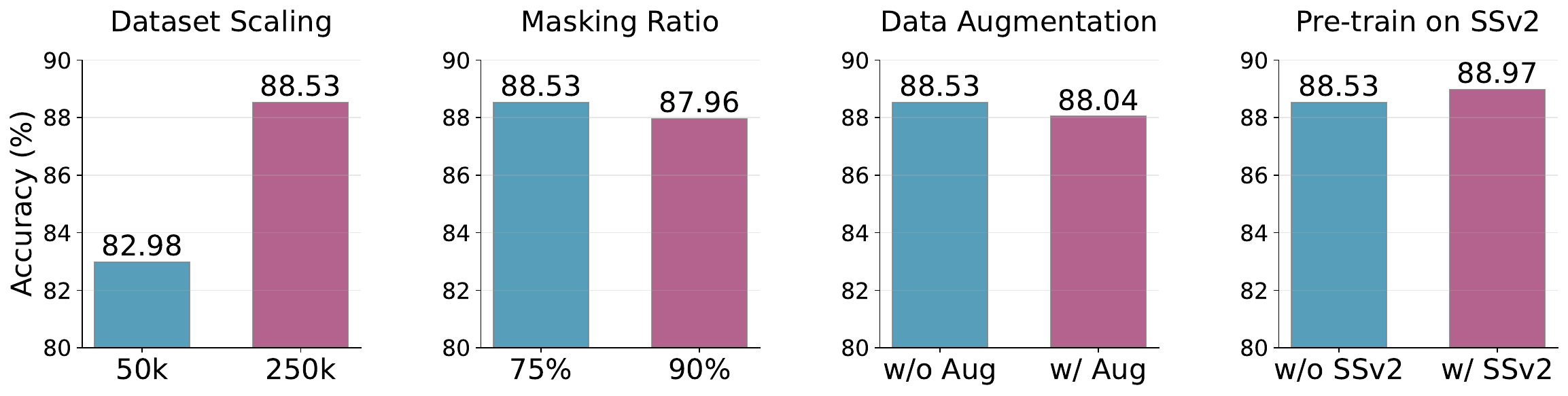}
    \caption{{\bf Ablation study of \emb~ on the ``Arrow of Time'' task.} We find that scaling the model from 50k to 250k Kubric samples leads to significant performance gains. Very high masking ratios (e.g., 90\% masking used in pixel-based video models) lead to worse performance likely due to point trajectories containing less redundant information than pixels. Simple data augmentation techniques (e.g., simulated camera zoom or panning effect via matrix operations on trajectory grids) do not lead to improved performance. We also experiment with a second stage pre-training on SSv2 (i.e., trajectories extracted using CoTracker3) for 30 epochs leading to a minor performance improvement.}
    \label{fig:ablation}
\end{figure}

We now measure the effect of different aspects of the model on the performance and show the results in Fig.~\ref{fig:ablation}. For this we use the ``Arrow of Time'' task on SSv2 from Sec.~\ref{subsec:temporal}. We observe that the results are rather stable with respect to most parameters. These include: varying the masking ratio during training time; augmenting data using synthetic camera motion (this is, not using the full Kubric pipeline, but simply adding panning to the point tracks); and fine-tuning on 50K videos of SSv2 on the self-supervised task. We observe that the masking ratio and data augmentation have a slight negative effect, while the pre-training on SSv2 have a slight positive effect. We also explore the effect of scaling data and observe that by far it has the strongest effect on performance. Multiplying the amount of training data by 5 yields a significant $5.4\%$ improvement. 
Additional ablation studies on token components, masking strategies, static camera simulations, and sim-to-real generalization are available in Appendix~\ref{appendix:additional_ablation_studies}.

%% file: tables/combined_video_understanding.tex
\begin{table}[t]
\centering
\caption{\textbf{Video Understanding Tasks. } We report top-1 accuracy across ~\emb~ features fused with Ego-Exo4D, SSv2, and Diving48. $\Delta$ denotes the improvement over the baseline. Standalone model performance can be found in Appendix~\ref{appendix:in_the_wild_appendix}. \\}
\label{tab:combined_video_understanding}
\resizebox{\textwidth}{!}{%
\begin{tabular}{@{}l cc cc cc@{}}
\toprule
& \multicolumn{2}{c}{\textbf{Ego-Exo4D}} & \multicolumn{2}{c}{\textbf{SSv2}} & \multicolumn{2}{c}{\textbf{Diving48}} \\
\cmidrule(lr){2-3} \cmidrule(lr){4-5} \cmidrule(lr){6-7}
\textbf{Model} & \textbf{Top-1 ($\pm$ Std \%)} & \textbf{$\Delta$} & \textbf{Top-1 ($\pm$ Std \%)} & \textbf{$\Delta$} & \textbf{Top-1 ($\pm$ Std \%)} & \textbf{$\Delta$} \\
\midrule
TIME + CLIP (4f) & 39.33 $\pm$ 0.24 & \textcolor{green!60!black}{$\mathbf{+14.22}$} & 31.56 $\pm$ 0.05 & \textcolor{green!60!black}{$\mathbf{+13.00}$} & 16.11 $\pm$ 0.28 & \textcolor{green!60!black}{$\mathbf{+2.14}$} \\
TIME + DINOv3 (4f) & 39.63 $\pm$ 0.22 & \textcolor{green!60!black}{$\mathbf{+14.75}$} & 34.60 $\pm$ 0.10 & \textcolor{green!60!black}{$\mathbf{+12.65}$} & 18.08 $\pm$ 0.21 & \textcolor{green!60!black}{$\mathbf{+1.73}$} \\
TIME + RVM$^\dagger$ & 37.00 $\pm$ 0.17 & \textcolor{green!60!black}{$\mathbf{+11.32}$} & 41.92 $\pm$ 0.06 & \textcolor{green!60!black}{$\mathbf{+1.84}$} & 11.74 $\pm$ 0.11 & \textcolor{green!60!black}{$\mathbf{+2.53}$} \\
TIME + VideoMAEv2$^\dagger$ & 51.14 $\pm$ 0.23 & \textcolor{green!60!black}{$\mathbf{+10.12}$} & 38.74 $\pm$ 0.09 & \textcolor{green!60!black}{$\mathbf{+5.69}$} & 24.97 $\pm$ 0.24 & \textcolor{green!60!black}{$\mathbf{+0.73}$} \\
TIME + V-JEPA 2$^\dagger$ & 45.60 $\pm$ 0.12 & \textcolor{green!60!black}{$\mathbf{+10.75}$} & 50.00 $\pm$ 0.10 & \textcolor{green!60!black}{$\mathbf{+0.22}$} & 30.04 $\pm$ 0.15 & \textcolor{gray}{$\mathbf{-0.04}$} \\
\bottomrule
\multicolumn{7}{l}{\footnotesize $^\dagger$ Denotes video models whose pre-training data explicitly includes the SSv2 dataset.} \\
\end{tabular}%
}
\end{table}

%% file: tables/temporal_tasks_table.tex
\begin{table}[t]
\centering
\caption{\textbf{Temporal Tasks}. Comparison of temporal directionality reasoning. Our purely kinematic model (\emb), trained exclusively on 140 hours of synthetic simulations, dramatically outperforms a size-matched RGB model (VideoMAEv2) and maintains high competitiveness against \vjepa, an architecture pre-trained on over $12,000$ times the real-world pre-training data.\\}
\label{tab:arrow_of_time}
\resizebox{1.0\textwidth}{!}{%
\begin{tabular}{l c c c >{\columncolor{lightblue}}c}
\toprule
 & VideoMAEv2$^\dagger$ & RVM$^\dagger$ & V-JEPA 2$^\dagger$ & TIME (Ours) \\
& \citep{wang2023videomaev2scalingvideo} & \citep{recurrentvideo} & \citep{vjepa2} & \\
\midrule
SSv2 (AoT) & 71.87 & 81.88 & \textbf{89.36} & \underline{88.53} \\
CLEVRER (T1$^\ast$) & $77.73 \pm 0.00$ & $86.56 \pm 1.17$ & $\underline{92.64 \pm 0.25}$ & $\mathbf{93.84 \pm 0.37}$ \\
CLEVRER (T2$^\ast$) & $33.33 \pm 0.00$ & $58.17 \pm 2.57$ & $\underline{58.92 \pm 3.40}$ & $\mathbf{74.95 \pm 7.06}$ \\
CLEVRER (T3$^\ast$) & $36.69 \pm 0.06$ & $59.10 \pm 0.68$ & $\mathbf{77.84 \pm 1.84}$ & \underline{$61.35 \pm 3.33$} \\
\midrule
Modality & 224$\times$224 Video & 224$\times$224 Video & 256$\times$256 Video & 32$\times$32 Tracks \\
Size & ViT-B & ViT-B & ViT-L & ViT-B \\
Dataset & Real-World & Real-World & Real-World & Synthetic \\
Training Samples & 5.36M ($27\times$) & 5.68M ($23\times$) & 22M ($88\times$) & 0.25M ($1\times$) \\
Video Hours Eq. & $\approx$7,500 ($54\times$) & $\approx$0.25M ($1,785\times$) & $\approx$1.73M ($12,357\times$) & 140 ($1\times$) \\
\bottomrule
\multicolumn{5}{l}{\footnotesize $^\dagger$ Denotes models whose pre-training data explicitly includes the SSv2 dataset.} \\
\multicolumn{5}{l}{\footnotesize $^\ast$ \textbf{T1}: Does a collision happen? \textbf{T2}: When does the first collision happen? \textbf{T3}: Number of collisions } \\
\end{tabular}%
}
\end{table}

%% file: limitations.tex

%% file: conclusion.tex
\section{Conclusion}
We have presented a self-supervised model called \emb~to learn video representations, based purely on temporal information in the form of point tracks. We have shown that learning video representation from motion alone has two core benefits. First, the model can learn to solve temporal tasks extremely well, on par or even surpassing state-of-the-art models trained on up to 4 orders of magnitude more data. Indeed, learning motion patterns proves to be extremely data-efficient, even compared to a very similar pixel-based architecture. Second, we have shown that this temporal model can be combined with appearance features, providing considerable improvements over using appearance alone on a wide range of tasks. We hope this model can be useful to the video research community either on its own for purely temporal tasks or as an add-on, easy to train model, that provides fundamental temporal information for perception.

%% file: tables/motion_only_baseline_results.tex

\begin{table}[h]
\centering
\caption{\textbf{Linear probe accuracy of trajectory-based methods on SSv2, Ego-Exo4D, and Diving48.} Each cell reports top-1 accuracy with standard deviation.\\}
\label{tab:motion_only_baseline_results}
\resizebox{\textwidth}{!}{%
\begin{tabular}{@{}l c c c@{}}
\toprule
\textbf{} & \textbf{SSv2} & \textbf{Ego-Exo4D} & \textbf{Diving48} \\
\textbf{Model} & \textbf{Top-1 ($\pm$ Std \%)} & \textbf{Top-1 ($\pm$ Std \%)} & \textbf{Top-1 ($\pm$ Std \%)} \\
\midrule
TIME (Ours) & \textbf{19.42} $\pm$ 0.04 & \textbf{17.27} $\pm$ 0.12 & \textbf{8.37} $\pm$ 0.15 \\
CoTracker3 backbone \citep{cotracker3} & 3.88 $\pm$ 0.02 & 10.42 $\pm$ 0.12 & \underline{6.48} $\pm$ 0.10 \\
TAPIR backbone \citep{doersch2023tapirtrackingpointperframe} & 4.77 $\pm$ 0.02 & \underline{12.33} $\pm$ 0.22 & 5.32 $\pm$ 0.23 \\
Features from raw trajectories & \underline{8.63} $\pm$ 0.02 & 2.87 $\pm$ 0.58 & 4.76 $\pm$ 0.28 \\
\bottomrule
\end{tabular}%
}
\end{table}

%% file: tables/masking_strategy_table.tex
\begin{table}[h]
\centering
\caption{\textbf{Comparison of TIME using first-frame versus middle-frame tracks on Something-Something v2.}}
\label{tab:masking_strategy_table}
\begin{tabular}{@{}lc@{}}
\toprule
\textbf{Baseline} & \textbf{Acc. (\%)} \\
\midrule
TIME (First frame tracks) & 19.42 $\pm$ 0.04 \\
TIME (Middle frame tracks) & 20.46 $\pm$ 0.12 \\
\bottomrule
\end{tabular}
\end{table}

%% file: tables/sim_to_real_generalization_table.tex
\begin{table}[h]
\centering
\caption{\textbf{Comparison of Kubric baseline and real-world trajectories on SSv2.} \\ }
\label{tab:sim_to_real_generalization_table}
\begin{tabular}{@{}lcc@{}}
\toprule
\textbf{Setting} & \textbf{Acc. (\%)} & \textbf{$\Delta$ (\%)} \\
\midrule
Baseline (Kubric) & 14.88 $\pm$ 0.08 & -- \\
Real-world traj. & 15.13 $\pm$ 0.03 & +1.68\% \\
\bottomrule
\end{tabular}
\end{table}

%% file: tables/static_camera_results.tex
\begin{table}[t]
\centering
\caption{\textbf{Accuracy (\%) on Something-Something v2 for selected camera motion labels.} Comparison between TIME and VideoMAEv2.\\}
\label{tab:static_camera_results}
\begin{tabular}{@{}l cc@{}}
\toprule
& \multicolumn{2}{c}{\textbf{Acc. (\%)}} \\
\textbf{Something-Something v2 Action Label} & \textbf{TIME} & \textbf{VideoMAEv2} \\
\midrule
Turning the camera right while filming something & 96.91 & 88.27 \\
Turning the camera left while filming something & 93.57 & 91.43 \\
Moving away from something with your camera & 84.09 & 69.32 \\
Approaching something with your camera & 83.62 & 78.45 \\
Turning the camera upwards while filming something & 66.67 & 76.67 \\
Turning the camera downwards while filming something & 62.90 & 67.74 \\
Moving something towards the camera & 47.30 & 45.95 \\
Moving something away from the camera & 44.64 & 32.14 \\
\bottomrule
\end{tabular}%
\end{table}

%% file: tables/token_information_ablation_table.tex
\begin{table}[h]
\centering
\caption{\textbf{Leave-one-out ablation study on TIME token components on Something-Something v2.}  \\}
\label{tab:token_information_ablation_table}

\begin{tabular}{@{}lcc@{}}
\toprule
\textbf{Setting} & \textbf{Acc. (\%)} & \textbf{$\Delta$ (\%)} \\
\midrule
Baseline (50k Kub.) & 14.88 $\pm$ 0.08 & -- \\
w/o KNN & 11.77 $\pm$ 0.10 & -20.90\% \\
w/o Global Disp. & 13.20 $\pm$ 0.16 & -11.29\% \\
w/o Occlusion & 11.36 $\pm$ 0.09 & -23.66\% \\
\bottomrule
\end{tabular}

\end{table}